\pdfoutput=1

\documentclass[11pt]{article}

\usepackage{emnlp2021}

\usepackage{times}
\usepackage{latexsym}

\usepackage[T1]{fontenc}

\usepackage[utf8]{inputenc}

\usepackage{microtype}

\usepackage{graphicx}
\usepackage{arydshln}
\usepackage[ruled,vlined]{algorithm2e}
\usepackage{algorithmic}
\usepackage{amsmath}
\usepackage{booktabs}
\newcommand{\kz}{\textit{K0}}
\newcommand{\ko}{\textit{K1}}
\newcommand{\kt}{\textit{K3}}
\newcommand{\dr}{\textit{d\&r}~}

\usepackage[colorinlistoftodos, textsize=tiny]{todonotes}
\usepackage{listings}
\definecolor{blued}{RGB}{70,197,221}

\title{Divide and Rule: Effective Pre-Training for Context-Aware Multi-Encoder Translation Models}

\author{
    Lorenzo Lupo$^1$ \: Marco Dinarelli$^1$ \: Laurent Besacier$^{1,2}$ \\
    $^1$Université Grenoble Alpes, France \\
    $^2$Naver Labs Europe, France \\
    \texttt{lorenzo.lupo@univ-grenoble-alpes.fr}\\
    \texttt{marco.dinarelli@univ-grenoble-alpes.fr}\\
    \texttt{laurent.besacier@naverlabs.com}
    }

\begin{document}
\maketitle
\begin{abstract}
Multi-encoder models are a broad family of context-aware neural machine translation systems that aim to improve translation quality by encoding document-level contextual information alongside the current sentence. The context encoding is undertaken by \textit{contextual parameters}, trained on document-level data. In this work, we discuss the difficulty of training these parameters effectively, due to the sparsity of the words in need of context (i.e., the training signal), and their relevant context. We propose to pre-train the contextual parameters over split sentence pairs, which makes an efficient use of the available data for two reasons. Firstly, it increases the contextual training signal by breaking intra-sentential syntactic relations, and thus pushing the model to search the context for disambiguating clues more frequently. Secondly, it eases the retrieval of relevant context, since context segments become shorter.
We propose four different splitting methods, and evaluate our approach with BLEU and contrastive test sets. Results show that it consistently improves learning of contextual parameters, both in low and high resource settings.
\end{abstract}

\section{Introduction}

Neural machine translation (NMT) has seen substantial improvements in recent years, fostered by the advent of the Transformer model~\citep{vaswani_attention_2017}. 
A remaining challenge for modern machine translation (MT) is the ability to contextualize translation of the current sentence with other sentences in the document~\citep{laubli_has_2018}. 
For this reason, contextual NMT has recently triggered a lot of attention and many approaches have been proposed in the literature. A common taxonomy~\citep{kim_when_2019,li_does_2020} divides them in two broad categories: single-encoder (concatenation) approaches~\citep{tiedemann_neural_2017, agrawal_contextual_2018, ma_simple_2020, zhang_long-short_2020} and multi-encoder approaches~\citep{jean_does_2017, tu_context_2017,bawden_evaluating_2018,miculicich_document-level_2018,voita_context-aware_2018,maruf_selective_2019,zheng_towards_2020}. Multi-encoder models are more flexible and can be more efficient than concatenation approaches, but they have been criticized as being mere regularization methods~\citep{kim_when_2019, li_does_2020}. In some cases, they have even been shown to perform worse than sentence-level systems on discourse-aware targeted test suites~\citep{lopes_document-level_2020}. 

In this work, we address this criticism by showing that training multi-encoder models is difficult because of two reasons: (i) the sparsity of \textit{contextual training signal}, i.e. the signal that pushes systems to translate in a context-aware fashion, which comes from the words that need context to be correctly translated; (ii) the sparsity of relevant context words, the ones needed to disambiguate translation. A trivial way to improve context-aware learning is by increasing the amount of document-level training data. Large document-level parallel corpora are not always available, but some works have proposed data augmentation techniques to remedy scarcity \cite{sugiyama_data_2019,stojanovski_contracat_2020,huo_diving_2020}. However, as we will show in our experimental section, this solution is not efficient and often sub-optimal. We therefore introduce a novel pre-training strategy, \textit{divide and rule (d\&r)}, that is based on a simple and yet powerful technique to augment the contextual training signal and to ease learning efficiently: splitting parallel sentences in segments (see Figure~\ref{fig:examplesplit}). Simply put, feeding a context-aware model with a sequence of incomplete, shorter, consecutive segments, forces it to look for context (i.e., surrounding segments) more frequently, and makes it easier to retrieve relevant context because segments are shorter. This results in faster and improved learning. 
We pre-train multi-encoder models on split datasets and evaluate them in two ways: BLEU score, and 
contrastive test sets for discourse phenomena.
\begin{figure}[t]
    \centering
    \resizebox{\columnwidth}{!}{%
    \begin{tabular}{ll}
    $S^{i,1}$ & He said that it was \underline{a project} of peace \\
    $S^{i,2}$ & and unity and that \underline{it} brought people together . \\
    $T^{i,1}$ & \textit{Il disait que c' était \underline{un projet} de paix} \\
    $T^{i,2}$ & \textit{et d' unité et qu' \underline{il} réunissait les gens .} \\
    \hdashline
    $S^{j,1}$ & I think single-cell \underline{organisms} \textbf{are} \\
    $S^{j,2}$ & \underline{possible} within two years . \\ 
    $T^{j,1}$ & \textit{Je pense que \underline{les organismes} unicellulaires} \\
    $T^{j,2}$ & \textit{\textbf{sont} \underline{possibles} dans 2 ans .}
    \end{tabular}
    }
    \caption{Example of sentence pairs from En$\rightarrow$Fr IWSLT17, after being tokenized and split in the middle. After the splitting, some syntactic relations span across two segments  (\underline{underlined}). Also, some source-side words are not parallel with their reference (\textbf{in bold}).}
    \label{fig:examplesplit}
\end{figure}

Our main contributions are the following: (i) we show that context-aware multi-encoder models need to be trained carefully, because the contextual training signal is sparse, as well as the context elements useful for contextualization;
(ii) we propose the \dr pre-training strategy, which fosters training of contextual parameters by splitting sentences into segments, with four splitting variants; (iii) we support this strategy with an analysis of the impact of splitting on the distribution of discourse phenomena; (iv) we demonstrate that this strategy is both effective and efficient, as it allows multi-encoder models to learn better and faster than by simply increasing the training data.
\section{Background}
\subsection{Single-encoder approaches}\label{subsec:singleenc}

The most straightforward approach to context-aware NMT consists in concatenating the 
context to the current sentence before feeding it to the standard encoder-decoder architecture~\citep{tiedemann_neural_2017, agrawal_contextual_2018, junczys-dowmunt_microsoft_2019, ma_simple_2020, zhang_long-short_2020}. A special token is introduced to 
mark the boundaries between sentences. Generation can then follow two strategies: the \textit{many-to-many} strategy consists in 
translating all the source sentences, and then discarding contextual sentences; the \textit{many-to-one} strategy consists in translating the current sentence only.
The modeling capacity of concatenation methods is limited to few sentences because the complexity of attention scales quadratically with sentence length, although some recent works try to solve this constraint \cite{tay_efficient_2020}.

\subsection{Multi-encoder approaches}\label{subsec:multienc}

\begin{figure}
    \centering
    \includegraphics[width=0.4\textwidth]{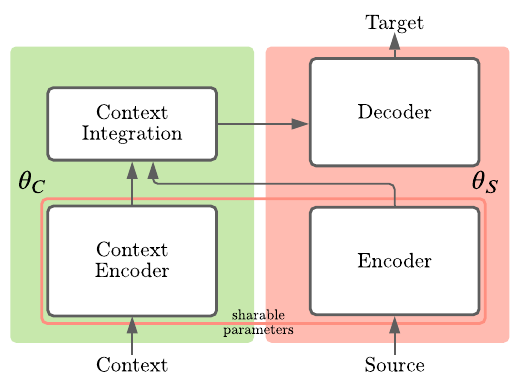}
    \caption{Multi-encoder  approach  integrating  context outside the decoder.}
    \label{fig:multiout}
\end{figure}

Multi-encoder models couple a self-standing sentence-level NMT system, with parameters $\theta_S$, with additional parameters 
for modeling the context either on source side, target side, or both. We refer to these parameters as the \textit{contextual parameters} $\theta_C$. The full context-aware architecture has parameters $\Theta=[\theta_S;\theta_C]$. Most of the multi-encoder models can be described as instances of two architectural families \cite{kim_when_2019}, that only differ in the way the encoded representations of the context and the current sentence are integrated.

\textbf{Outside integration.} In this approach, depicted in Figure~\ref{fig:multiout}, the encoded representations are merged outside the decoder~\citep{maruf_contextual_2018, voita_context-aware_2018,zhang_improving_2018,miculicich_document-level_2018,maruf_selective_2019,zheng_towards_2020}. This can happen in different ways, such as by simple concatenation of the encodings, or with a gated sum.

\textbf{Inside integration.} Here the decoder attends to the context representations  directly, using its internal representation of the decoded history as query~\citep{tu_learning_2018, kuang_modeling_2018, bawden_evaluating_2018, voita_when_2019, tan_hierarchical_2019}.

Many of these works found it useful to share parameters 
of current-sentence and context encoders~\citep{voita_context-aware_2018, li_does_2020}. In this way, the amount of contextual parameters to learn, $|\theta_C|$, and the computational cost are drastically reduced. 
Shared representation can also be cached to be re-used and further processed by contextual parameters without the need of re-encoding  sentences from scratch, which represents an advantage with respect to single-encoder approaches. 
Most of the approaches proposed in the literature focus on a few previous sentences, where most of the relevant context is concentrated.

\textbf{Two-step training.} Multi-encoder models are commonly trained following a two-step strategy \cite{tu_learning_2018,zhang_improving_2018,miculicich_document-level_2018,maruf_document_2018,li_does_2020}. The first step consists in training $\theta_S$ independently on a sentence-level parallel corpus $\mathcal{C}_S$. Secondarily, contextual parameters $\theta_C$ are trained on a document-level parallel corpus $\mathcal{C}_D$, while fine-tuning or freezing $\theta_S$. Note that $\mathcal{C}_S$ can also include sentences from $\mathcal{C}_D$.

\subsection{Evaluating context-aware MT}\label{subsec:evaluation}

Novel MT systems are usually evaluated by computing BLEU \cite{papineni_bleu_2002} on the test data. However, BLEU is ill-equipped to capture the improvements achieved by context-aware MT~\citep{hardmeier_discourse_2012}, because contextualization can improve the translation of only a small fraction of the words in a document, while most of the words can be correctly translated without knowing the context. For instance, only a fraction of the anaphoric pronouns in a document has its nominal antecedent outside its own sentence. However, despite being sparse, these few cases strongly impact the quality of translation  ~\citep{laubli_has_2018,popescu-belis_context_2019}. Consequently, a number of discourse-targeted test sets and automatic metrics have been proposed to measure improvements in context-aware MT \cite{maruf_survey_2019}, the most widely adopted ones being contrastive test sets.

\textbf{Contrastive test sets} 
\citep{bawden_evaluating_2018,muller_large-scale_2018,voita_context-aware_2019} consist of a number of source sentences, each paired with a correct translation and some incorrect ones. Models are assessed on their ability to rank the correct translation first. 
In many cases, 
this can be identified only by looking at context, which is provided for both source and target sides. Therefore, the ranking accuracy reflects the context-modeling ability of the evaluated translation system.

\section{The double challenge of sparsity }\label{sec:sparsesignal}

Some works criticized multi-encoder methods ~\citep{kim_when_2019,li_does_2020}, arguing that they do not improve sentence-level baselines in terms of BLEU when the baseline is well regularized. When there are improvements, it is argued that the context-encoder simply works as a noise-generator that makes training more robust, and the improvements are not 
due to better context-modeling. Along this path, \citet{lopes_document-level_2020} showed that multi-encoder architectures struggle to model contextual information, and even deteriorate the performance of a sentence-level baseline on contrastive test sets. In fact, many proponents of multi-encoder models only show BLEU improvements, without providing any kind of targeted evaluation. 
This doesn't allow a direct evaluation of their context-modeling capability. 
We posit that training the contextual parameters of multi-encoder models is non-trivial because of two challenges: (i) the sparsity of the training signal, which comes from the words that need context to be correctly translated (most of the words of a sentence can be translated without context); 
(ii) the sparsity of context words that are useful for contextualization (most of the context is useless). As such, missing the right experimental setting can lead to unsuccessful training and unconvincing results.

\textbf{More data?} A trivial way to offset sparsity is to increase the volume of training data. In fact, existing works that report strong results with targeted evaluation train their contextual parameters with millions of document-level sentence pairs ~\citep{bawden_evaluating_2018, muller_large-scale_2018,voita_when_2019,zheng_towards_2020,wong_contextual_2020,kang_dynamic_2020}. In contrast, many works in the literature train models with the TED talks' subtitles released by the IWSLT shared tasks~\citep{cettolo_wit3_2012}, which only consist of a couple of hundred thousand parallel sentences. In the experimental section, we will show that IWSLT's subtitles are not sufficient to effectively train multi-encoder models. It follows that one can not make fair comparisons between alternative architectures in such experimental settings. On the other hand, we will give an empirical confirmation to the intuition that increasing the volume of training data helps learning contextual parameters.
However, increasing the amount of training data is an inefficient solution, and one that is not always feasible: large document-level training sets may not be available in many languages. In the following section, we propose a pre-training solution that makes an efficient use of the available data for learning contextual-parameters effectively.

\section{Proposed Approach}\label{sec:proposedapproach}

\begin{algorithm}[t]
    \SetAlgoLined
    \caption{Split parallel corpus}
	\begin{algorithmic}[1]
	\STATE {\bfseries input:} Parallel corpus $\mathcal{C}$, minimum source length $l_{min}$, function $\mathrm{wheresplit}()$
	\FOR{$i=1,\dots,|\mathcal{C}|$}
	    \IF{$len(S^i)\geq l_{min}$}
		    \STATE $m_S, m_T=\mathrm{wheresplit}(S^i, T^i, ...)$
		    \STATE $S^{i,1}=S^i_{< m_S}$ and $S^{i,2}=S^i_{\geq m_S}$
		    \STATE $T^{i,1}=T^i_{< m_T}$ and $T^{i,2}=T^i_{\geq m_T}$
		\ENDIF
	\ENDFOR
	\RETURN Split corpus $\mathcal{C}_D$
	\end{algorithmic}
	\label{alg:split}
\end{algorithm}

One way to simulate 
document-level data is to split sentences in two or more segments \cite{luong_multi-task_2016}.
In this way intra-sentential syntactic relations are broken, and a word previously disambiguated by looking at its neighbours in the sentence, now requires contextual information from the other segment in order to be correctly translated. Moreover, splitting sentences increases the concentration of relevant words within the context segment, as we will show in Section \ref{subsec:analysis}.
Within the framework of MT, if we split the source sentence, its corresponding reference has to be split too. The proposed approach, \textit{divide and rule} (\dr), consists in pre-training the model on a dataset $\mathcal{C}_D$ that results from splitting all the sentences of a parallel corpus $\mathcal{C}$ that have at least $l_{min}$ tokens, as described by Algorithm~\ref{alg:split}. Each source-side sentence $S^i$, with index $i=1,...,|\mathcal{C}|$, is split into $S^{i,1}$ and $S^{i,2}$. Its corresponding reference $T^i$ is split into $T^{i,1}$ and $T^{i,2}$. The resulting corpus is a document-level parallel corpus $\mathcal{C}_D$, such that, if the original corpus $C$ was itself document-level, then $\mathcal{C}_D$ keeps the same document boundaries as $C$.
Figure~\ref{fig:examplesplit} illustrates two examples of parallel sentences that are split in the middle. In both examples, a context-aware system needs to look at $S^{i,1}$ for translating $S^{i,2}$ correctly, i.e. to look at past context. In the first one, the English neuter pronoun ``it" could be translated into ``il" or ``elle", according to the gender of its antecedent (there is no singular neuter 3rd-person in French). The antecedent ``a project", which is in the previous segment, allows to disambiguate it into ``il". In the second example, the adjective ``possible'' can be correctly translated into its plural version ``possibles'' by looking back at the noun it refers to: ``organisms''.



\subsection{Splitting methods}\label{subsec:splitmethod}

In Algoritm~\ref{alg:split}, the $\mathrm{wheresplit}$ function returns the token indices $m_S$ and $m_T$ of $S^i$ and $T^i$, where the sentence is split.
In this work, we propose and experiment with four variants of this function.

\textbf{Middle-split.} The simplest strategy is to split both the source and the target in the middle. In this case, $\mathrm{wheresplit}=\mathrm{middlesplit}(S^i, T^i)$ returns $m_S=\lfloor len(S^i)/2\rfloor$ and $m_T=\lfloor len(T^i)/2\rfloor$. Following this method, it can happen that $S^{i,j}$ and $T^{i,j}$, with $j=1,2$, are not parallel, as illustrated in the second example of Figure~\ref{fig:examplesplit}. The verb ``are'' belongs to $S^{i,1}$, but its translation ``sont'' does not belong to its corresponding reference segment $T^{i,1}$. In other words, sometimes the splitting can separate a set of words from their reference, which end up in the other segment. Clearly, this method requires the two languages not to excessively diverge in terms of word order, to avoid too large mismatches between $S^{i,j}$ and $T^{i,j}$, with $j=1,2$.

\textbf{Aligned-split.}
As a solution to the misalignment problem between source and target segments, we can calculate word alignments $A^i$, and use them to inform our splitting strategy by setting $\mathrm{wheresplit}=\mathrm{alignedsplit}(S^i, T^i, A^i)$, where  $\mathrm{alignedsplit}$ splits each sentence close to the middle, while avoiding to separate aligned words in different segments.

\textbf{Synt-split}. The objective of splitting being to break intra-sentential syntactic and semantic relations in order to force the model to exploit the context more frequently, we can 
run an NLP toolkit over the training set to retrieve 
relations $L$ (e.g. syntactic dependencies or coreferences), and then by defining $\mathrm{wheresplit}=\mathrm{syntsplit}(S^i, T^i, L^i)$ so that it splits sentences close to the middle, while ensuring that at least a relation is broken whenever possible. Since not all relations raise translation ambiguities when broken, one can choose which of them must be prioritized; in this work we chose pronominal coreferences. 

\textbf{Multi-split.} The aforementioned methods can be extended to splitting sentences in more than two segments. The more we split sentences the more likely it is that context is needed for each segment, hence increasing training signal for contextual parameters.

In Section~\ref{subsec:splitmethodexp}, we present an empirical comparison between the four splitting methods. We refer to Appendix~\ref{app:methods} and to our implementation\footnote{https://github.com/lorelupo/divide-and-rule} for further details.

\subsection{Impact on discourse phenomena}\label{subsec:analysis}

\begin{figure}
    \centering
    \includegraphics[scale=0.42]{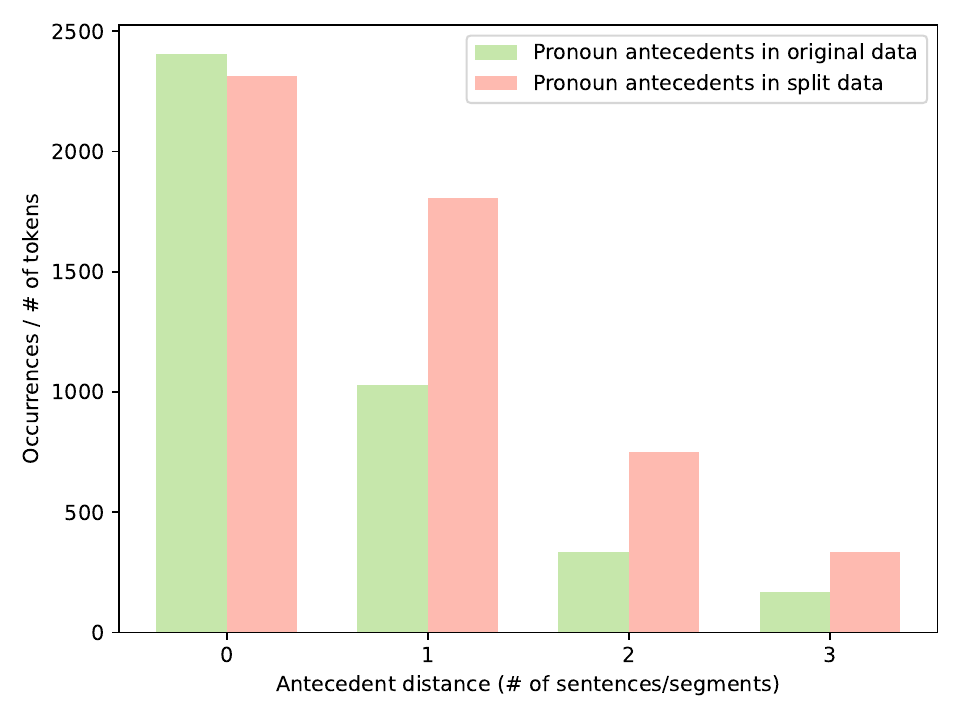}
    \caption{IWSLT's number of antecedents of anaphoric pronouns at a given distance in terms of sentences or segments (in split data), divided by the number of tokens that the model needs to attend for resolving the coreference.}
    \label{fig:normsplitcoref}
\end{figure}

To give an explicit picture of how and why sentence splitting helps learning contextual parameters, we processed the source training data of IWSLT17 with CoreNLP \cite{manning_stanford_2014} and we computed some statistics on coreference chains and dependency parse trees, before and after applying the \textit{middle-split} method. Statistics show how splitting the sentences of a document helps in two ways:

\textbf{More cases.} Splitting generates new cases that require context for disambiguation, making training signal more abundant. When syntactic dependencies are split in two segments, the model needs to access the context for reconstructing the syntactic structure of the source sentence and correctly translate it, as shown in Figure~\ref{fig:examplesplit}. In order to have an idea of the magnitude of this effect, we calculated the percentage of the sentences where the splitting method breaks at least one syntactic dependency between the main verb of the sentence (the root) and : (i) the subject or object (18.1\% of the sentences); (ii) any complement (9.5\%); (iii) any modifier (9.3\%). If we consider all the dependencies with the root, except punctuations, we find that in 84.8\% of the sentences at least a syntactic dependency is broken.
Given such high proportion, the \textit{middle-split} variant is in fact a good approximation of a syntactically supported splitting approach. These cases add up to the many other cases of broken relations, such as coreferences, which make the overall contextual training signal more abundant. 

\textbf{Denser cases.} The splitting also has the effect of shortening the average length of text sequences, which eases the job of context-aware systems because they have to attend to fewer words while looking for context. In Figure~\ref{fig:normsplitcoref}, we show how many antecedents of an anaphoric pronoun are present in the data at a given distance $d$, expressed as number of sentences from the current one for original data, and number of segments for split data. $d=0$ means that both the pronoun and its antecedent are in the same sentence (or segment); $d=1$ means that the antecedent is in previous sentence (or segment), and so on. We show statistics up to $d=3$, which is the maximum context distance that we experiment with. The absolute number of antecedents is divided by the average length of a sentence or segment. The resulting bar plot shows that splitting sentences into segments makes pronominal antecedents more dense in the set of context tokens that the model is attending, which fosters the learning of contextual parameters. The same effect applies to the other discourse phenomena that require contextual disambiguation.\footnote{More details are available in Appendix~\ref{app:impact}, along with the same statistics for Opensubtitles2018.}

\section{Experimental setup}\label{sec:setup}

\subsection{Data}

We conduct experiments for three language pairs: English$\rightarrow$Russian/German/French, on different domains. Following \citet{kim_when_2019}, we pre-train sentence-level baselines on large sentence-level parallel data to make them as robust as possible. In particular, we employ data released by \citet{voita_when_2019} for En$\rightarrow$Ru (6.0M sentences from OpenSubtitles2018 \citep{lison_opensubtitles2018_2018}), data from the WMT17\footnote{http://www.statmt.org/wmt17/translation-task.html} news translation shared task for En$\rightarrow$De ($\sim$5.2M sentences), and data from WMT14\footnote{http://www.statmt.org/wmt14/translation-task.html} for En$\rightarrow$Fr ($\sim$35.8M sentences).
We train the contextual parameters of context-aware models in two settings, while freezing the rest of the parameters:

\textbf{High resource data.} For En$\rightarrow$Ru, it consists of the document-level data released by \citet{voita_when_2019}. For the other two language pairs, we build the training set by assembling (i) News-Commentary-v12 for En$\rightarrow$De and News-Commentary-v9 for En$\rightarrow$Fr; (ii) Europarl-v7 for En$\rightarrow$De/Fr; (iii) TED talks subtitles released by IWSLT17~\cite{cettolo_wit3_2012} for En$\rightarrow$De/Fr.

\textbf{Low resource data.} For En$\rightarrow$Ru, it consists of a random subset of the high resource documents, amounting to 1/10th of its total. For En$\rightarrow$De/Fr, we use IWSLT17's TED talks alone.

The resulting size of the two training settings after pre-processing is reported in Table~\ref{tab:datastats}. In the case of En$\rightarrow$De/Fr, baselines and context-aware models that were trained on high resources are also fine-tuned on IWSLT17, so that both high and low resource settings can be bench-marked on the IWSLT17's test set 2015. Test-sets 2011-2014 are used as development set. For En$\rightarrow$Ru, we use the dev and test sets provided by \citet{voita_when_2019}. \footnote{We report in Appendix~\ref{app:setting} a re-cap of the datasets used and details about pre-processing.}

\begin{table}[t!]
\centering
\small
\begin{tabular}{llll}
\toprule
 & En$\rightarrow$Ru & En$\rightarrow$De & En$\rightarrow$Fr \\
\midrule
Low Res & 0.15M (8.3) & 0.20M (20.8) & 0.23M (21.0) \\
High Res & 1.50M (8.3) & 2.29M (27.29) & 2.31M (27.6) \\
\bottomrule
\end{tabular}
\caption{Millions of sentence pairs used for training context-aware models, and their average source length.}
\label{tab:datastats}
\end{table}

\subsection{Evaluation}
Besides evaluating average translation quality with BLEU~\citep{papineni_bleu_2002},\footnote{Moses' \textit{multi-bleu-detok}~\citep{koehn_moses_2007} for De/Fr, \textit{multi-bleu} on lowercased Ru as~\citet{voita_when_2019}.} we employ three contrastive test suites for the evaluation of the translation of discourse phenomena.\footnote{Whenever relevant, we calculate the statistical significance of the differences between models' accuracies with the paired McNemar test~\citep{mcnemar_note_1947}.}

\textbf{En$\rightarrow$Ru EllipsisVP}~\citep{voita_when_2019}. Consisting of 500 examples from  OpenSubtitles2018, each containing multiple contrastive hypotheses to evaluate the translation of verb phrase ellipses.  Source sentences contain an auxiliary verb (e.g. "do") and an omitted main verb, which can be imputed thanks to one of the three context sentences. \citet{voita_when_2019} proposed test sets for the evaluation of other discourse phenomena, but we do not use them because they are conceived for systems using target-side context too. 

\textbf{En$\rightarrow$De ContraPro}~\citep{muller_large-scale_2018}. A large-scale test set from  OpenSubtitles2018~\citep{lison_opensubtitles2018_2018}, that measures translation accuracy of the English anaphoric pronoun \textit{it} into the corresponding German translations \textit{er}, \textit{sie} or \textit{es}. Examples are balanced across the three pronoun classes (4,000 examples each). Each example requires identification of the pronominal antecedent, either in the source or target side, that can be found in the current sentence or any of the previous ones.

\textbf{En$\rightarrow$Fr ContraPro}~\citep{lopes_document-level_2020}. A large-scale test set from  OpenSubtitles2018, completely analogous to the previous one but focused on the translation of two English pronouns: \textit{it} and \textit{they}. It consists of 3,500 examples for each target pronoun type: \textit{il} or \textit{elle} for \textit{it}, \textit{ils} or \textit{elles} for \textit{they}.


\subsection{Models}

We experiment with three models:

\textbf{\kz.} A sentence-level baseline, following the \textit{Transformer-base} by \citet{vaswani_attention_2017}.

\textbf{\ko.} A context aware multi-encoder architecture with \textit{outside integration} (see Section~\ref{subsec:multienc}), that encodes a single past source sentence as context.

\textbf{\kt.} A context aware multi-encoder architecture with \textit{outside integration}, that encodes three past source sentences as context.\footnote{Although the splitting does not increase the number of inter-segment phenomena for $d>1$, it strengthens the signal by making it more dense (see Section~\ref{subsec:analysis}). Thus, \kt{} and any wider-context model can profit from the proposed approach.}

For both \ko\ and \kt, sentence-level parameters $\theta_S$ follow the \textit{Transformer-base} configuration (hidden size of 512, feed forward size of  2048,  6  layers,  8  attention  heads, total of 60.7M parameters), while contextual parameters $\theta_C$ follow hierarchical architecture with source-side encoder proposed by \citet{miculicich_document-level_2018} (hidden size of 512, feed forward size of  2048, 8  attention heads, total of 4.7M parameters).\footnote{Details can be found in Appendix~\ref{app:setting}} Context-aware models are trained following the \textit{two-step strategy}  described in Section~\ref{subsec:multienc}.  Sentence-level parameters $\theta_S$ of both \ko\ and \kt\ are initialized with \kz\ 
and freezed.
This has the advantage of saving time and computation, since only a small fraction of parameters ($\theta_C$) is trained (4.7M over a total of 65.2M).

\section{Results and Analysis}\label{sec:results}
\subsection{Training contextual parameters is hard}\label{subsec:lowres}

\begin{table*}[t!]
\centering
\small
\begin{tabular}{lc|cc|cc|cc|c}
\toprule
\textbf{} & \textbf{} & \multicolumn{2}{c}{En$\rightarrow$De} & \multicolumn{2}{c}{En$\rightarrow$Fr} & \multicolumn{2}{c}{En$\rightarrow$Ru} & Avg. \\
Model & Setting & BLEU & \textbf{ContraPro$\uparrow$} & BLEU & \textbf{ContraPro$\uparrow$} & BLEU & \textbf{Ellipsis-VP$\uparrow$} & Train Hours \\
\midrule
\textit{K0} & - & 32.97 & 46.37 & 41.63 & 79.46 & 31.37 & 25.40 & - \\
\hdashline\textit{K1} & Low Res & 33.14 & 47.05 & 41.93 & 79.24 & 30.89 & 32.20 & 2.9 (1.0x) \\
\textit{K3} & Low Res & 32.86 & 46.48 & 41.40 & 80.53 & 31.00 & 29.20 & 3.5 (1.0x) \\
\textit{K1} & High Res & 33.16 & 57.75 & 41.65 & 84.32 & 31.15 & 44.00 & 13.0 (4.5x) \\
\textit{K3} & High Res & 33.1 & 51.14 & 41.95 & 82.94 & 31.23 & 39.20 & 16.8 (4.8x) \\
\midrule
\textit{K1-d\&r} & Low Res & 33.44 & 60.21* & 41.78 & 84.06 & 31.09 & 47.00* & 6.7 (2.3x) \\
\textit{K3-d\&r} & Low Res & 33.36 & 56.22* & 41.68 & 85.50* & 32.12 & 46.60* & 6.4 (1.8x) \\
\textit{K1-d\&r} & High Res & 32.82 & \textbf{61.09*} & 41.81 & 84.17 & 31.09 & 59.40* & 16.5 (5.7x) \\
\textit{K3-d\&r} & High Res & 33.07 & 59.56* & 41.91 & \textbf{85.66*} & 31.27 & \textbf{60.40*} & 22.3 (6.4x) \\
\bottomrule
\end{tabular}
\caption{BLEU score on testsets and accuracy (\%) on contrastive sets. The last column reports the average context-aware training time (in hours), including the time for \dr pre-training. The symbol * denotes statistically significant (p<0.01) improvements w.r.t \kz{} and non-\dr counterparts  (first block of rows). }
\label{tab:main}
\end{table*}

In this section we provide evidence about the difficulty of training contextual parameters on document-level data. In the first block of lines of Table~\ref{tab:main}, after the results of the sentence-level baseline \kz, we report  performance of context-aware models trained on original document-level data, comparing low and high resource settings. 

\textbf{When trained on low resources}, models display good BLEU on the test set, generally without relevant degradation with respect to \kz{}, or even with some improvements. However, such marginal fluctuations in BLEU are difficult to interpret, as they do not necessarily correspond to better or worse translation~\citep{freitag_bleu_2020}. Accuracy on the contrastive test sets also increases marginally over baseline, if at all, for En$\rightarrow$De/Fr. \ko{} even shows a slight degradation of performance over the sentence-level baseline for En$\rightarrow$Fr. These results highlight the struggle of contextual parameters to learn an appropriate use of context, other than acting as mere regularizers, as it was suggested by \citet{kim_when_2019} and~\citet{li_does_2020}. On Russian instead, models display some improvements w.r.t. \kz. This aligns with our expectations, since En$\rightarrow$Ru Low Res has a volume of inter-sentential discourse phenomena (such as coreferences)  that is comparable with En$\rightarrow$De/Fr Low Res, but sentences are 2.5x shorter.\footnote{See Table~\ref{tab:datastats}; more details can be found in Appendix~\ref{app:impact}} In other words, the \textit{double challenge of sparsity} is mitigated on this corpus. 

\textbf{When trained on high resources}, systems show substantial improvements in their context-modeling capabilities, on all language pairs. Instead, BLEU improves of a few decimal points only, showing its limits to measure improvements in context-aware translation.  These results confirm the intuition discussed in Section \ref{sec:sparsesignal}: increasing the volume of data is a trivial solution to mitigate sparsity. 

\subsection{Divide and rule}

In this section, we show that the proposed pre-training strategy is a more efficient answer to the double challenge of sparsity than simply adding more data, and one that allows improvements when resources are abundant too. The second block of Table~\ref{tab:main} displays performance of models that have undergone \dr pre-training on the same document-level data as the models in the previous block, but where sentences were split in two segments following the \textit{middle-split} method with $l_{min}=7$ (see \ref{subsec:splitmethod}). During \dr pre-training, \ko{} and \kt{} encode one and three past segments (instead of sentences), respectively.
After \dr pre-training, models have been tuned and evaluated on original, non-split data. The pre-training proves to be very effective, as all models show strong improvements in terms of accuracy on the test suites, with the sole exception of \textit{K1-\dr} on En$\rightarrow$Fr High Res. The average improvement is of \textbf{+10.79} accuracy points on Low Res, \textbf{+8.49} on High Res, showing that \dr brings strong improvements even when data are abundant. Interestingly, improvements are not uniformly distributed across language pairs and domains: \textbf{+17.20} on average for En$\rightarrow$Ru, \textbf{+8.67} for En$\rightarrow$De, \textbf{+3.09} for En$\rightarrow$Fr. In terms of BLEU instead, we keep seeing minor fluctuations. This confirms that, while context-aware translation improves dramatically, the average translation quality measured with BLEU stays more or less constant.\footnote{To verify that the improvements on test suites after \dr pre-training really come from a better use of context, we present in Appendix~\ref{app:bydistance} an analysis of pronoun translation by antecedent distance, and an ablation study in which we test models on ContraPro with inconsistent context.}
It is now evident that a proper assessment of multi-encoder approaches can not be undergone without careful training of contextual parameters that targets the problem of sparsity.

\textbf{Efficiency}. A comparison between -\dr models trained on Low Res against models trained on High Res without \textit{d\&r} shows another quality of the \dr pre-training strategy: efficiency. The same context-aware models achieve superior performances with 1/10th of the document-level data and a much shorter training time (last column).

\subsection{Impact of the splitting method}\label{subsec:splitmethodexp}

\begin{table}[t]
\centering
\small
\begin{tabular}{lcccc}
\toprule
 \multicolumn{5}{c}{En$\rightarrow$De} \\
 \midrule
 & Middle$\uparrow$ & Aligned$\uparrow$ & Synt$\uparrow$ & Multi$\uparrow$ \\
\midrule
\textit{K1-d\&r} & 60.21 & +0.69* & -2.67* & - \\
\textit{K3-d\&r} & 56.22 & -1.38* & +1.33* & +1.13* \\
\midrule
 \multicolumn{5}{c}{En$\rightarrow$Fr} \\
 \midrule
\textit{K1-d\&r} & 84.06 & +0.27 & +0.15 & - \\
\textit{K3-d\&r} & 85.50 & +0.20 & +0.33** & -0.09 \\
\bottomrule
\end{tabular}
\caption{Comparison of accuracy of context-aware pronoun translation (ContraPro) by \dr pre-trained models with the \textit{middle-split} method (first column) and the other proposed methods (relative difference). *: $p<0.01$, **: $p<0.05$.}
\label{tab:splitmethod}
\end{table}

Following Section~\ref{subsec:splitmethod}, we study the impact of using a different splitting method other than \textit{middle-split}. All the variants are applied to the En$\rightarrow$De/Fr low resource setting (IWSLT), with $l_{min}=7$, and the \dr pre-trained models are evaluated on ContraPro. The \textit{aligned-split} method is based on alignments learned with \textit{fast\_align}~\citep{dyer_simple_2013}, while for the \textit{synt-split} method we retrieve intra-sentential pronominal coreferences with CoreNLP \cite{manning_stanford_2014}, and we try to split them wherever present in a sentence-pair. We split sentences as close to the middle as possible, while attempting to break the maximum number of coreferences.\footnote{More sophisticated \textit{synt-split} methods could be devised, targeting other discourse phenomena, or several of them at the same time, with different degrees of priority.}  Finally, for the \textit{multi-split} method, we split sentence-pairs in a half for $len(S^i)\geq 7$, and also in three segments of identical size for $len(S^i)\geq 15$. The performance differences between models pre-trained with \textit{middle-split} and the other variants are reported in Table~\ref{tab:splitmethod}. As we can see, splitting variants allow small improvements in 7 cases out of 10, although variations are marginal: the simple \textit{middle-split} method seems to be already close to optimal. This observation can be explained by multiple elements. Firstly, \textit{middle-split} produces segment pairs that are already well aligned: most of the source and target segments are aligned with the exception of one or two words, and the fact of having only a few misplaced words might act as a regularization factor. Secondly, \textit{middle-split} breaks a syntactic relation for the vast majority of sentences already, as explained in Section~\ref{subsec:analysis}, which means that improvements achieved with syntactically driven splitting can only be marginal. Thirdly, splitting in more than one segment can be beneficial in some cases, because it allows to break more syntactic relations and increase density of signal, but it also increases the risk of misalignment between source and target, and might make the task too hard. Finally, tools like \textit{fast\_align} and CoreNLP are characterized by a non-negligible language-dependent error rate, which affects the performance of the methods. In conclusion, \dr pre-training with \textit{middle-split} seems to be the most convenient alternative for most use-cases because of its efficacy, its simplicity and its language-independence.

\subsubsection{On the scope of middle-split}\label{app:scope}

Even though \textit{middle-split} relies on word order similarity between source and target languages, we argue that the required degree of similarity is met by a large number of language pairs, in the order of millions. In fact, there are around 4,000 written languages in the world~\cite{eberhard_ethnologue_2021}, and most of them can be grouped in a few types with similar word orders, as shown by the ample literature on word order typologies~\cite{tomlin_basic_2014,dryer_world_2013}. \\
The primary order of interest is the \textit{constituent order}, concerning the relative order of subject (S), object (O) and verb (V) in a clause. There are seven possible language types with respect to the constituent order~\cite{dryer_order_2013}: SOV, SVO, VSO, VOS, OVS, OSV, NDO (non-dominant order). \citet{tomlin_basic_2014} estimates that more than 40\% of the world languages belong to the SOV type (languages adopting the SOV order), another 40\% belong to the SVO type, while almost 10\% of languages adopt VSO order. The other types are rarer. As we have shown above, the \textit{middle-split} method is beneficial both in the case of language pairs of the same type, that deploy the same constituent order, like En-Fr/Ru, which all adopt SVO order, as well as for language pairs that belong to different types, as for En-De, where English is SVO and German is NDO, deploying both SOV and SVO according to the use cases~\cite{dryer_order_2013}.\\
Similar observations also apply when we look at other word order categories. For instance, when looking at the order of modifiers or adverbials, languages can be clustered in a few types too, where the wide majority of languages belong to the biggest or second biggest type \cite{dryer_order_2013-1,dryer_order_2013-2}. Therefore, we believe that our method can be beneficial for millions of language pairs, including many low resource languages belonging not only to same word order types, but also slightly different ones (as in the case of SOV and SVO).

\subsubsection{Benchmarking}

\begin{table}[]
\centering
\small
\begin{tabular}{lcccc}
\toprule
\textbf{}           & \multicolumn{2}{c}{En$\rightarrow$De} & \multicolumn{2}{c}{En$\rightarrow$Fr} \\
Model               & BLEU   & \textbf{ContraP} & BLEU       & \textbf{ContraP}      \\\midrule
K0                  & 32.08  & 45.00                        & 40.92      & 79.70                    \\\hdashline
\textit{Zhang2018}  & 31.03  & 42.60                        & 40.95      & 59.00                \\
\textit{Tu2018*} & 32.10  & 45.20                        & 40.91      & 79.70                \\
\textit{Concat21} & 31.84  & 48.00                        & 40.67      & 80.90                \\
\textit{Concat22*} & 30.89  & \textbf{70.80}                        & 40.57      & 83.20                \\\midrule
\textit{K1-d\&r}    & \textbf{33.44}  & 60.21               & \textbf{41.78}      & 84.06                \\
\textit{K3-d\&r}    & 33.36  & 56.22                        & 41.68      & \textbf{85.50} \\\bottomrule  
\end{tabular}
\caption{Comparison between various context-aware models trained on low resources (1st block) and  our \dr pre-trained multi-encoder models trained on the same data. We report BLEU scores on the IWSLT testset and accuracies (in \%) on  ContraPro. The models annotated with * have the advantage of accessing target-side context, while all the other context aware models access source-side context only.}
\label{tab:benchmark}
\end{table}

For a wider contextualization of our results, we report in the first block of Table~\ref{tab:benchmark} some experimental results by \citet{lopes_document-level_2020}, who trained and evaluated various context-aware approaches on the same low-resource setting (IWSLT17), adopting the very same experimental setup as ours.
Specifically, they trained and evaluated:

\textbf{\kz{}}: a baseline like ours;

\textbf{\textit{Zhang2018}}: a multi-encoder model that encodes three past source sentences, with both \textit{inside integration} (see \ref{subsec:multienc}) and \textit{outside integration}~\citep{zhang_improving_2018};

\textbf{\textit{Tu2018}}: a multi-encoder model that encodes all the past context (at any distance) with a caching system with \textit{inside integration} in the decoder, both on the source and the target side~\citep{tu_learning_2018};

\textbf{Concat21}: a \textit{many-to-one} (2-to-1) single-encoder approach (see \ref{subsec:singleenc}) that exploits contextual information from one past source sentence;

\textbf{Concat22}: a \textit{many-to-many} (2-to-2) single-encoder approach that exploits contextual information from one past sentence, both on the source and the target side;

Multi-encoder models (\textit{Zhang2018} and \textit{Tu2018}) perform poorly or even lag behind \kz, confirming the difficulty of multi-encoder models to learn contextualization on low resources and without any help against the problem of sparsity. Instead, concatenation approaches are stronger, likely because they do not have extra contextual parameters to train and simply finetune the same sentence-level Transformer architecture on the context-aware task. This makes them less affected by the problem of sparsity. Our splitting strategy proves to be very effective, since both \dr pre-trained models outperform \textit{Concat21} using the same amount of training data. Moreover, \textit{K1/3}-\dr beat the strong \textit{Concat22} on En$\rightarrow$Fr, which has the non-negligible advantage of using also the target-side context along the source-side. This benchmarking show that multi-encoder models are a viable solution for context-aware NMT, but that they need to be carefully (pre)-trained to harness their capabilities. We leave to future works a more detailed comparison between single-encoder and multi-encoder approaches, as well as between \dr and other recently proposed pre-training strategies for context-aware models (e.g., \citet{fernandes_measuring_2021}).


\section{Conclusions}

Multi-encoder models are a broad family of context-aware NMT models. In this work we have discussed the difficulty of training contextual parameters due to the sparsity of the words in need of context, and the sparsity of their relevant context. We have proposed a pre-training approach called \textit{divide and rule}, based on splitting the training sentences, with four variants. After having analysed the implications of splitting on the distribution of discourse phenomena within the training data, we have shown that \dr allows to learn contextual parameters better and faster than by simply increasing training data. We have also shown that the simplest and language independent splitting variant, \textit{middle-split}, is a strong baseline that can be easily applied for pre-training any multi-encoder NMT model.

\section*{Acknowledgements}
We thank the anonymous reviewers for their insightful comments.
This work has been partially supported by the Multidisciplinary Institute in Artificial Intelligence MIAI@Grenoble Alpes (ANR-19-P3IA-0003), and it was granted access to the HPC resources of IDRIS under the allocation 2020-101501 made by GENCI.

\bibliographystyle{acl_natbib}
\bibliography{document_nmt}

\begin{thebibliography}{56}
\expandafter\ifx\csname natexlab\endcsname\relax\def\natexlab#1{#1}\fi

\bibitem[{Agrawal et~al.(2018)Agrawal, Turchi, and
  Negri}]{agrawal_contextual_2018}
Ruchit~Rajeshkumar Agrawal, Marco Turchi, and Matteo Negri. 2018.
\newblock \href {https://cris.fbk.eu/handle/11582/314425#.XlZ3xeF7mkA}
  {Contextual {Handling} in {Neural} {Machine} {Translation}: {Look} {Behind},
  {Ahead} and on {Both} {Sides}}.
\newblock In \emph{Proceedings of the 21st {Annual} {Conference} of the
  {European} {Association} for {Machine} {Translation}}, pages 11--20, Alacant,
  Spain.

\bibitem[{Bawden et~al.(2018)Bawden, Sennrich, Birch, and
  Haddow}]{bawden_evaluating_2018}
Rachel Bawden, Rico Sennrich, Alexandra Birch, and Barry Haddow. 2018.
\newblock \href {https://doi.org/10.18653/v1/N18-1118} {Evaluating discourse
  phenomena in neural machine translation}.
\newblock In \emph{Proceedings of the 2018 Conference of the North {A}merican
  Chapter of the Association for Computational Linguistics: Human Language
  Technologies, Volume 1 (Long Papers)}, pages 1304--1313, New Orleans,
  Louisiana. Association for Computational Linguistics.

\bibitem[{Cettolo et~al.(2012)Cettolo, Girardi, and
  Federico}]{cettolo_wit3_2012}
Mauro Cettolo, Christian Girardi, and Marcello Federico. 2012.
\newblock \href {https://www.aclweb.org/anthology/2012.eamt-1.60} {{WIT}3: Web
  inventory of transcribed and translated talks}.
\newblock In \emph{Proceedings of the 16th Annual conference of the European
  Association for Machine Translation}, pages 261--268, Trento, Italy. European
  Association for Machine Translation.

\bibitem[{Dryer(2013{\natexlab{a}})}]{dryer_order_2013-2}
Matthew~S. Dryer. 2013{\natexlab{a}}.
\newblock \href {https://ci.nii.ac.jp/naid/10030056551/} {Order of {Adjective}
  and {Noun}}.
\newblock In \emph{The {World} {Atlas} of {Language} {Structures} {Online}}.
  Max Planck Institute for Evolutionary Anthropology, Leipzig, Germany.

\bibitem[{Dryer(2013{\natexlab{b}})}]{dryer_order_2013-1}
Matthew~S. Dryer. 2013{\natexlab{b}}.
\newblock \href {https://ci.nii.ac.jp/naid/10030056551/} {Order of {Adverbial}
  {Subordinator} and {Clause}}.
\newblock In \emph{The {World} {Atlas} of {Language} {Structures} {Online}}.
  Max Planck Institute for Evolutionary Anthropology, Leipzig, Germany.

\bibitem[{Dryer(2013{\natexlab{c}})}]{dryer_order_2013}
Matthew~S. Dryer. 2013{\natexlab{c}}.
\newblock \href {https://ci.nii.ac.jp/naid/10030056551/} {Order of subject,
  object and verb}.
\newblock In \emph{The {World} {Atlas} of {Language} {Structures} {Online}}.
  Max Planck Institute for Evolutionary Anthropology, Leipzig, Germany.

\bibitem[{Dryer and Haspelmath(2013)}]{dryer_world_2013}
Matthew~S. Dryer and Martin Haspelmath. 2013.
\newblock \href {http://wals.info/} {\emph{The {World} {Atlas} of {Language}
  {Structures} {Online}}}.
\newblock Max Planck Institute for Evolutionary Anthropology, Leipzig, Germany.

\bibitem[{Dyer et~al.(2013)Dyer, Chahuneau, and Smith}]{dyer_simple_2013}
Chris Dyer, Victor Chahuneau, and Noah~A. Smith. 2013.
\newblock \href {https://www.aclweb.org/anthology/N13-1073} {A simple, fast,
  and effective reparameterization of {IBM} model 2}.
\newblock In \emph{Proceedings of the 2013 Conference of the North {A}merican
  Chapter of the Association for Computational Linguistics: Human Language
  Technologies}, pages 644--648, Atlanta, Georgia. Association for
  Computational Linguistics.

\bibitem[{Eberhard et~al.(2021)Eberhard, Simons, and
  Fenning}]{eberhard_ethnologue_2021}
David~M. Eberhard, Gary~F. Simons, and Charles~D. Fenning. 2021.
\newblock \href {https://www.ethnologue.com/} {Ethnologue: {Languages} of the
  {World}}.
\newblock Library Catalog: www.ethnologue.com.

\bibitem[{Fernandes et~al.(2021)Fernandes, Yin, Neubig, and
  Martins}]{fernandes_measuring_2021}
Patrick Fernandes, Kayo Yin, Graham Neubig, and Andr{\'e} F.~T. Martins. 2021.
\newblock \href {https://doi.org/10.18653/v1/2021.acl-long.505} {Measuring and
  increasing context usage in context-aware machine translation}.
\newblock In \emph{Proceedings of the 59th Annual Meeting of the Association
  for Computational Linguistics and the 11th International Joint Conference on
  Natural Language Processing (Volume 1: Long Papers)}, pages 6467--6478,
  Online. Association for Computational Linguistics.

\bibitem[{Freitag et~al.(2020)Freitag, Grangier, and
  Caswell}]{freitag_bleu_2020}
Markus Freitag, David Grangier, and Isaac Caswell. 2020.
\newblock \href {https://doi.org/10.18653/v1/2020.emnlp-main.5} {{BLEU} might
  be guilty but references are not innocent}.
\newblock In \emph{Proceedings of the 2020 Conference on Empirical Methods in
  Natural Language Processing (EMNLP)}, pages 61--71, Online. Association for
  Computational Linguistics.

\bibitem[{Hardmeier(2012)}]{hardmeier_discourse_2012}
Christian Hardmeier. 2012.
\newblock \href {https://doi.org/10.4000/discours.8726} {Discourse in
  {Statistical} {Machine} {Translation}. {A} {Survey} and a {Case} {Study}}.
\newblock \emph{Discours. Revue de linguistique, psycholinguistique et
  informatique. A journal of linguistics, psycholinguistics and computational
  linguistics}, (11).
\newblock 00039 Number: 11 Publisher: Presses universitaires de Caen.

\bibitem[{Huo et~al.(2020)Huo, Herold, Gao, Dahlmann, Khadivi, and
  Ney}]{huo_diving_2020}
Jingjing Huo, Christian Herold, Yingbo Gao, Leonard Dahlmann, Shahram Khadivi,
  and Hermann Ney. 2020.
\newblock \href {https://www.aclweb.org/anthology/2020.wmt-1.71} {Diving deep
  into context-aware neural machine translation}.
\newblock In \emph{Proceedings of the Fifth Conference on Machine Translation},
  pages 604--616, Online. Association for Computational Linguistics.

\bibitem[{Jean et~al.(2017)Jean, Lauly, Firat, and Cho}]{jean_does_2017}
Sebastien Jean, Stanislas Lauly, Orhan Firat, and Kyunghyun Cho. 2017.
\newblock \href {http://arxiv.org/abs/1704.05135} {Does {Neural} {Machine}
  {Translation} {Benefit} from {Larger} {Context}?}
\newblock \emph{arXiv:1704.05135 [cs, stat]}.
\newblock 00039 arXiv: 1704.05135.

\bibitem[{Junczys-Dowmunt(2019)}]{junczys-dowmunt_microsoft_2019}
Marcin Junczys-Dowmunt. 2019.
\newblock \href {https://doi.org/10.18653/v1/W19-5321} {{M}icrosoft translator
  at {WMT} 2019: Towards large-scale document-level neural machine
  translation}.
\newblock In \emph{Proceedings of the Fourth Conference on Machine Translation
  (Volume 2: Shared Task Papers, Day 1)}, pages 225--233, Florence, Italy.
  Association for Computational Linguistics.

\bibitem[{Kang et~al.(2020)Kang, Zhao, Zhang, and Zong}]{kang_dynamic_2020}
Xiaomian Kang, Yang Zhao, Jiajun Zhang, and Chengqing Zong. 2020.
\newblock \href {https://doi.org/10.18653/v1/2020.emnlp-main.175} {Dynamic
  context selection for document-level neural machine translation via
  reinforcement learning}.
\newblock In \emph{Proceedings of the 2020 Conference on Empirical Methods in
  Natural Language Processing (EMNLP)}, pages 2242--2254, Online. Association
  for Computational Linguistics.

\bibitem[{Kim et~al.(2019)Kim, Tran, and Ney}]{kim_when_2019}
Yunsu Kim, Duc~Thanh Tran, and Hermann Ney. 2019.
\newblock \href {https://doi.org/10.18653/v1/D19-6503} {When and why is
  document-level context useful in neural machine translation?}
\newblock In \emph{Proceedings of the Fourth Workshop on Discourse in Machine
  Translation (DiscoMT 2019)}, pages 24--34, Hong Kong, China. Association for
  Computational Linguistics.

\bibitem[{Koehn et~al.(2007)Koehn, Hoang, Birch, Callison-Burch, Federico,
  Bertoldi, Cowan, Shen, Moran, Zens, Dyer, Bojar, Constantin, and
  Herbst}]{koehn_moses_2007}
Philipp Koehn, Hieu Hoang, Alexandra Birch, Chris Callison-Burch, Marcello
  Federico, Nicola Bertoldi, Brooke Cowan, Wade Shen, Christine Moran, Richard
  Zens, Chris Dyer, Ond{\v{r}}ej Bojar, Alexandra Constantin, and Evan Herbst.
  2007.
\newblock \href {https://www.aclweb.org/anthology/P07-2045} {{M}oses: Open
  source toolkit for statistical machine translation}.
\newblock In \emph{Proceedings of the 45th Annual Meeting of the Association
  for Computational Linguistics Companion Volume Proceedings of the Demo and
  Poster Sessions}, pages 177--180, Prague, Czech Republic. Association for
  Computational Linguistics.

\bibitem[{Kuang et~al.(2018)Kuang, Xiong, Luo, and Zhou}]{kuang_modeling_2018}
Shaohui Kuang, Deyi Xiong, Weihua Luo, and Guodong Zhou. 2018.
\newblock \href {https://www.aclweb.org/anthology/C18-1050} {Modeling coherence
  for neural machine translation with dynamic and topic caches}.
\newblock In \emph{Proceedings of the 27th International Conference on
  Computational Linguistics}, pages 596--606, Santa Fe, New Mexico, USA.
  Association for Computational Linguistics.

\bibitem[{L{\"a}ubli et~al.(2018)L{\"a}ubli, Sennrich, and
  Volk}]{laubli_has_2018}
Samuel L{\"a}ubli, Rico Sennrich, and Martin Volk. 2018.
\newblock \href {https://doi.org/10.18653/v1/D18-1512} {Has machine translation
  achieved human parity? a case for document-level evaluation}.
\newblock In \emph{Proceedings of the 2018 Conference on Empirical Methods in
  Natural Language Processing}, pages 4791--4796, Brussels, Belgium.
  Association for Computational Linguistics.

\bibitem[{Li et~al.(2020)Li, Liu, Wang, Jiang, Xiao, Zhu, Liu, and
  Li}]{li_does_2020}
Bei Li, Hui Liu, Ziyang Wang, Yufan Jiang, Tong Xiao, Jingbo Zhu, Tongran Liu,
  and Changliang Li. 2020.
\newblock \href {https://doi.org/10.18653/v1/2020.acl-main.322} {Does
  multi-encoder help? a case study on context-aware neural machine
  translation}.
\newblock In \emph{Proceedings of the 58th Annual Meeting of the Association
  for Computational Linguistics}, pages 3512--3518, Online. Association for
  Computational Linguistics.

\bibitem[{Lison et~al.(2018)Lison, Tiedemann, and
  Kouylekov}]{lison_opensubtitles2018_2018}
Pierre Lison, J{\"o}rg Tiedemann, and Milen Kouylekov. 2018.
\newblock \href {https://www.aclweb.org/anthology/L18-1275}
  {{O}pen{S}ubtitles2018: Statistical rescoring of sentence alignments in
  large, noisy parallel corpora}.
\newblock In \emph{Proceedings of the Eleventh International Conference on
  Language Resources and Evaluation ({LREC} 2018)}, Miyazaki, Japan. European
  Language Resources Association (ELRA).

\bibitem[{Lopes et~al.(2020)Lopes, Farajian, Bawden, Zhang, and
  Martins}]{lopes_document-level_2020}
Ant{\'o}nio Lopes, M.~Amin Farajian, Rachel Bawden, Michael Zhang, and
  Andr{\'e} F.~T. Martins. 2020.
\newblock \href {https://www.aclweb.org/anthology/2020.eamt-1.24}
  {Document-level neural {MT}: A systematic comparison}.
\newblock In \emph{Proceedings of the 22nd Annual Conference of the European
  Association for Machine Translation}, pages 225--234, Lisboa, Portugal.
  European Association for Machine Translation.

\bibitem[{Luong et~al.(2016)Luong, Le, Sutskever, Vinyals, and
  Kaiser}]{luong_multi-task_2016}
Thang Luong, Quoc~V. Le, Ilya Sutskever, Oriol Vinyals, and Lukasz Kaiser.
  2016.
\newblock Multi-task sequence to sequence learning.
\newblock In \emph{International Conference on Learning Representations}.

\bibitem[{Ma et~al.(2020)Ma, Zhang, and Zhou}]{ma_simple_2020}
Shuming Ma, Dongdong Zhang, and Ming Zhou. 2020.
\newblock \href {https://doi.org/10.18653/v1/2020.acl-main.321} {A simple and
  effective unified encoder for document-level machine translation}.
\newblock In \emph{Proceedings of the 58th Annual Meeting of the Association
  for Computational Linguistics}, pages 3505--3511, Online. Association for
  Computational Linguistics.

\bibitem[{Manning et~al.(2014)Manning, Surdeanu, Bauer, Finkel, Bethard, and
  McClosky}]{manning_stanford_2014}
Christopher Manning, Mihai Surdeanu, John Bauer, Jenny Finkel, Steven Bethard,
  and David McClosky. 2014.
\newblock \href {https://doi.org/10.3115/v1/P14-5010} {The {S}tanford
  {C}ore{NLP} natural language processing toolkit}.
\newblock In \emph{Proceedings of 52nd Annual Meeting of the Association for
  Computational Linguistics: System Demonstrations}, pages 55--60, Baltimore,
  Maryland. Association for Computational Linguistics.

\bibitem[{Maruf and Haffari(2018)}]{maruf_document_2018}
Sameen Maruf and Gholamreza Haffari. 2018.
\newblock \href {https://doi.org/10.18653/v1/P18-1118} {Document context neural
  machine translation with memory networks}.
\newblock In \emph{Proceedings of the 56th Annual Meeting of the Association
  for Computational Linguistics (Volume 1: Long Papers)}, pages 1275--1284,
  Melbourne, Australia. Association for Computational Linguistics.

\bibitem[{Maruf et~al.(2018)Maruf, Martins, and
  Haffari}]{maruf_contextual_2018}
Sameen Maruf, Andr{\'e} F.~T. Martins, and Gholamreza Haffari. 2018.
\newblock \href {https://doi.org/10.18653/v1/W18-6311} {Contextual neural model
  for translating bilingual multi-speaker conversations}.
\newblock In \emph{Proceedings of the Third Conference on Machine Translation:
  Research Papers}, pages 101--112, Brussels, Belgium. Association for
  Computational Linguistics.

\bibitem[{Maruf et~al.(2019)Maruf, Martins, and Haffari}]{maruf_selective_2019}
Sameen Maruf, Andr{\'e} F.~T. Martins, and Gholamreza Haffari. 2019.
\newblock \href {https://doi.org/10.18653/v1/N19-1313} {Selective attention for
  context-aware neural machine translation}.
\newblock In \emph{Proceedings of the 2019 Conference of the North {A}merican
  Chapter of the Association for Computational Linguistics: Human Language
  Technologies, Volume 1 (Long and Short Papers)}, pages 3092--3102,
  Minneapolis, Minnesota. Association for Computational Linguistics.

\bibitem[{Maruf et~al.(2021)Maruf, Saleh, and Haffari}]{maruf_survey_2019}
Sameen Maruf, Fahimeh Saleh, and Gholamreza Haffari. 2021.
\newblock \href {https://doi.org/10.1145/3441691} {A survey on document-level
  neural machine translation: Methods and evaluation}.
\newblock \emph{ACM Comput. Surv.}, 54(2).

\bibitem[{McNemar(1947)}]{mcnemar_note_1947}
Quinn McNemar. 1947.
\newblock \href {https://doi.org/10.1007/BF02295996} {Note on the sampling
  error of the difference between correlated proportions or percentages}.
\newblock \emph{Psychometrika}, 12(2):153--157.
\newblock 03511.

\bibitem[{Miculicich et~al.(2018)Miculicich, Ram, Pappas, and
  Henderson}]{miculicich_document-level_2018}
Lesly Miculicich, Dhananjay Ram, Nikolaos Pappas, and James Henderson. 2018.
\newblock \href {https://doi.org/10.18653/v1/D18-1325} {Document-level neural
  machine translation with hierarchical attention networks}.
\newblock In \emph{Proceedings of the 2018 Conference on Empirical Methods in
  Natural Language Processing}, pages 2947--2954, Brussels, Belgium.
  Association for Computational Linguistics.

\bibitem[{M{\"u}ller et~al.(2018)M{\"u}ller, Rios, Voita, and
  Sennrich}]{muller_large-scale_2018}
Mathias M{\"u}ller, Annette Rios, Elena Voita, and Rico Sennrich. 2018.
\newblock \href {https://doi.org/10.18653/v1/W18-6307} {A large-scale test set
  for the evaluation of context-aware pronoun translation in neural machine
  translation}.
\newblock In \emph{Proceedings of the Third Conference on Machine Translation:
  Research Papers}, pages 61--72, Brussels, Belgium. Association for
  Computational Linguistics.

\bibitem[{Ott et~al.(2019)Ott, Edunov, Baevski, Fan, Gross, Ng, Grangier, and
  Auli}]{ott_fairseq_2019}
Myle Ott, Sergey Edunov, Alexei Baevski, Angela Fan, Sam Gross, Nathan Ng,
  David Grangier, and Michael Auli. 2019.
\newblock \href {https://doi.org/10.18653/v1/N19-4009} {fairseq: A fast,
  extensible toolkit for sequence modeling}.
\newblock In \emph{Proceedings of the 2019 Conference of the North {A}merican
  Chapter of the Association for Computational Linguistics (Demonstrations)},
  pages 48--53, Minneapolis, Minnesota. Association for Computational
  Linguistics.

\bibitem[{Papineni et~al.(2002)Papineni, Roukos, Ward, and
  Zhu}]{papineni_bleu_2002}
Kishore Papineni, Salim Roukos, Todd Ward, and Wei-Jing Zhu. 2002.
\newblock \href {https://doi.org/10.3115/1073083.1073135} {{B}leu: a method for
  automatic evaluation of machine translation}.
\newblock In \emph{Proceedings of the 40th Annual Meeting of the Association
  for Computational Linguistics}, pages 311--318, Philadelphia, Pennsylvania,
  USA. Association for Computational Linguistics.

\bibitem[{Pereyra et~al.(2017)Pereyra, Tucker, Chorowski, Kaiser, and
  Hinton}]{Pereyra_regularizing_2017}
Gabriel Pereyra, George Tucker, Jan Chorowski, Lukasz Kaiser, and Geoffrey
  Hinton. 2017.
\newblock \href {http://arxiv.org/abs/1701.06548} {Regularizing {Neural}
  {Networks} by {Penalizing} {Confident} {Output} {Distributions}}.
\newblock \emph{ICLR workshop}.
\newblock 00464 arXiv: 1701.06548.

\bibitem[{Popel and Bojar(2018)}]{popel_training_2018}
Martin Popel and Ondřej Bojar. 2018.
\newblock \href {https://doi.org/10.2478/pralin-2018-0002} {Training {Tips} for
  the {Transformer} {Model}}.
\newblock \emph{The Prague Bulletin of Mathematical Linguistics},
  110(1):43--70.

\bibitem[{Popescu-Belis(2019)}]{popescu-belis_context_2019}
Andrei Popescu-Belis. 2019.
\newblock \href {http://arxiv.org/abs/1901.09115} {Context in {Neural}
  {Machine} {Translation}: {A} {Review} of {Models} and {Evaluations}}.
\newblock \emph{arXiv:1901.09115 [cs]}.
\newblock 00010 arXiv: 1901.09115.

\bibitem[{Scherrer et~al.(2019)Scherrer, Tiedemann, and
  Lo{\'a}iciga}]{scherrer_analysing_2019}
Yves Scherrer, J{\"o}rg Tiedemann, and Sharid Lo{\'a}iciga. 2019.
\newblock \href {https://doi.org/10.18653/v1/D19-6506} {Analysing concatenation
  approaches to document-level {NMT} in two different domains}.
\newblock In \emph{Proceedings of the Fourth Workshop on Discourse in Machine
  Translation (DiscoMT 2019)}, pages 51--61, Hong Kong, China. Association for
  Computational Linguistics.

\bibitem[{Sennrich et~al.(2016)Sennrich, Haddow, and
  Birch}]{sennrich_neural_2016}
Rico Sennrich, Barry Haddow, and Alexandra Birch. 2016.
\newblock \href {https://doi.org/10.18653/v1/P16-1162} {Neural machine
  translation of rare words with subword units}.
\newblock In \emph{Proceedings of the 54th Annual Meeting of the Association
  for Computational Linguistics (Volume 1: Long Papers)}, pages 1715--1725,
  Berlin, Germany. Association for Computational Linguistics.

\bibitem[{Stojanovski et~al.(2020)Stojanovski, Krojer, Peskov, and
  Fraser}]{stojanovski_contracat_2020}
Dario Stojanovski, Benno Krojer, Denis Peskov, and Alexander Fraser. 2020.
\newblock \href {https://doi.org/10.18653/v1/2020.coling-main.417}
  {{C}ontra{CAT}: Contrastive coreference analytical templates for machine
  translation}.
\newblock In \emph{Proceedings of the 28th International Conference on
  Computational Linguistics}, pages 4732--4749, Barcelona, Spain (Online).
  International Committee on Computational Linguistics.

\bibitem[{Sugiyama and Yoshinaga(2019)}]{sugiyama_data_2019}
Amane Sugiyama and Naoki Yoshinaga. 2019.
\newblock \href {https://doi.org/10.18653/v1/D19-6504} {Data augmentation using
  back-translation for context-aware neural machine translation}.
\newblock In \emph{Proceedings of the Fourth Workshop on Discourse in Machine
  Translation (DiscoMT 2019)}, pages 35--44, Hong Kong, China. Association for
  Computational Linguistics.

\bibitem[{Tan et~al.(2019)Tan, Zhang, Xiong, and Zhou}]{tan_hierarchical_2019}
Xin Tan, Longyin Zhang, Deyi Xiong, and Guodong Zhou. 2019.
\newblock \href {https://doi.org/10.18653/v1/D19-1168} {Hierarchical modeling
  of global context for document-level neural machine translation}.
\newblock In \emph{Proceedings of the 2019 Conference on Empirical Methods in
  Natural Language Processing and the 9th International Joint Conference on
  Natural Language Processing (EMNLP-IJCNLP)}, pages 1576--1585, Hong Kong,
  China. Association for Computational Linguistics.

\bibitem[{Tay et~al.(2020)Tay, Dehghani, Bahri, and
  Metzler}]{tay_efficient_2020}
Yi~Tay, Mostafa Dehghani, Dara Bahri, and Donald Metzler. 2020.
\newblock \href {http://arxiv.org/abs/2009.06732} {Efficient {Transformers}:
  {A} {Survey}}.
\newblock \emph{arXiv:2009.06732 [cs]}.
\newblock 00008 arXiv: 2009.06732.

\bibitem[{Tiedemann and Scherrer(2017)}]{tiedemann_neural_2017}
J{\"o}rg Tiedemann and Yves Scherrer. 2017.
\newblock \href {https://doi.org/10.18653/v1/W17-4811} {Neural machine
  translation with extended context}.
\newblock In \emph{Proceedings of the Third Workshop on Discourse in Machine
  Translation}, pages 82--92, Copenhagen, Denmark. Association for
  Computational Linguistics.

\bibitem[{Tomlin(2014)}]{tomlin_basic_2014}
Russell~S. Tomlin. 2014.
\newblock \emph{Basic {Word} {Order} ({RLE} {Linguistics} {B}: {Grammar}):
  {Functional} {Principles}}.
\newblock Routledge.
\newblock Google-Books-ID: OlPIAgAAQBAJ.

\bibitem[{Tu et~al.(2017)Tu, Liu, Lu, Liu, and Li}]{tu_context_2017}
Zhaopeng Tu, Yang Liu, Zhengdong Lu, Xiaohua Liu, and Hang Li. 2017.
\newblock \href {https://doi.org/10.1162/tacl_a_00048} {Context gates for
  neural machine translation}.
\newblock \emph{Transactions of the Association for Computational Linguistics},
  5:87--99.

\bibitem[{Tu et~al.(2018)Tu, Liu, Shi, and Zhang}]{tu_learning_2018}
Zhaopeng Tu, Yang Liu, Shuming Shi, and Tong Zhang. 2018.
\newblock \href {https://doi.org/10.1162/tacl_a_00029} {Learning to remember
  translation history with a continuous cache}.
\newblock \emph{Transactions of the Association for Computational Linguistics},
  6:407--420.

\bibitem[{Vaswani et~al.(2017)Vaswani, Shazeer, Parmar, Uszkoreit, Jones,
  Gomez, Kaiser, and Polosukhin}]{vaswani_attention_2017}
Ashish Vaswani, Noam Shazeer, Niki Parmar, Jakob Uszkoreit, Llion Jones,
  Aidan~N. Gomez, Lukasz Kaiser, and Illia Polosukhin. 2017.
\newblock Attention is all you need.
\newblock In \emph{Proceedings of the 31st {International} {Conference} on
  {Neural} {Information} {Processing} {Systems}}, {NIPS}'17, pages 6000--6010,
  Long Beach, California, USA. Curran Associates Inc.

\bibitem[{Voita et~al.(2019{\natexlab{a}})Voita, Sennrich, and
  Titov}]{voita_context-aware_2019}
Elena Voita, Rico Sennrich, and Ivan Titov. 2019{\natexlab{a}}.
\newblock \href {https://doi.org/10.18653/v1/D19-1081} {Context-aware
  monolingual repair for neural machine translation}.
\newblock In \emph{Proceedings of the 2019 Conference on Empirical Methods in
  Natural Language Processing and the 9th International Joint Conference on
  Natural Language Processing (EMNLP-IJCNLP)}, pages 877--886, Hong Kong,
  China. Association for Computational Linguistics.

\bibitem[{Voita et~al.(2019{\natexlab{b}})Voita, Sennrich, and
  Titov}]{voita_when_2019}
Elena Voita, Rico Sennrich, and Ivan Titov. 2019{\natexlab{b}}.
\newblock \href {https://doi.org/10.18653/v1/P19-1116} {When a good translation
  is wrong in context: Context-aware machine translation improves on deixis,
  ellipsis, and lexical cohesion}.
\newblock In \emph{Proceedings of the 57th Annual Meeting of the Association
  for Computational Linguistics}, pages 1198--1212, Florence, Italy.
  Association for Computational Linguistics.

\bibitem[{Voita et~al.(2018)Voita, Serdyukov, Sennrich, and
  Titov}]{voita_context-aware_2018}
Elena Voita, Pavel Serdyukov, Rico Sennrich, and Ivan Titov. 2018.
\newblock \href {https://doi.org/10.18653/v1/P18-1117} {Context-aware neural
  machine translation learns anaphora resolution}.
\newblock In \emph{Proceedings of the 56th Annual Meeting of the Association
  for Computational Linguistics (Volume 1: Long Papers)}, pages 1264--1274,
  Melbourne, Australia. Association for Computational Linguistics.

\bibitem[{Wong et~al.(2020)Wong, Maruf, and Haffari}]{wong_contextual_2020}
KayYen Wong, Sameen Maruf, and Gholamreza Haffari. 2020.
\newblock \href {https://doi.org/10.18653/v1/2020.acl-main.530} {Contextual
  neural machine translation improves translation of cataphoric pronouns}.
\newblock In \emph{Proceedings of the 58th Annual Meeting of the Association
  for Computational Linguistics}, pages 5971--5978, Online. Association for
  Computational Linguistics.

\bibitem[{Zhang et~al.(2018)Zhang, Luan, Sun, Zhai, Xu, Zhang, and
  Liu}]{zhang_improving_2018}
Jiacheng Zhang, Huanbo Luan, Maosong Sun, Feifei Zhai, Jingfang Xu, Min Zhang,
  and Yang Liu. 2018.
\newblock \href {https://doi.org/10.18653/v1/D18-1049} {Improving the
  transformer translation model with document-level context}.
\newblock In \emph{Proceedings of the 2018 Conference on Empirical Methods in
  Natural Language Processing}, pages 533--542, Brussels, Belgium. Association
  for Computational Linguistics.

\bibitem[{Zhang et~al.(2020)Zhang, Chen, Ge, and Fan}]{zhang_long-short_2020}
Pei Zhang, Boxing Chen, Niyu Ge, and Kai Fan. 2020.
\newblock \href {https://doi.org/10.18653/v1/2020.emnlp-main.81} {Long-short
  term masking transformer: A simple but effective baseline for document-level
  neural machine translation}.
\newblock In \emph{Proceedings of the 2020 Conference on Empirical Methods in
  Natural Language Processing (EMNLP)}, pages 1081--1087, Online. Association
  for Computational Linguistics.

\bibitem[{Zheng et~al.(2020)Zheng, Yue, Huang, Chen, and
  Birch}]{zheng_towards_2020}
Zaixiang Zheng, Xiang Yue, Shujian Huang, Jiajun Chen, and Alexandra Birch.
  2020.
\newblock \href {https://doi.org/10.24963/ijcai.2020/551} {Towards {Making} the
  {Most} of {Context} in {Neural} {Machine} {Translation}}.
\newblock In \emph{Proceedings of the {Twenty}-{Ninth} {International} {Joint}
  {Conference} on {Artificial} {Intelligence}, {IJCAI}-20}, pages 3983--3989.
  International Joint Conferences on Artificial Intelligence Organization.

\end{thebibliography}

\appendix


\begin{table}[ht]
    \centering
    \small
    \begin{tabular}{lrrr}
    \toprule
    \multicolumn{4}{c}{Coreferences - original data} \\
    \midrule
    $d$    &   \#tokens    &   \multicolumn{2}{c}{Occurrences} \\
    &   &   All    &   Pronouns \\
    \midrule
    $0$     &   21.01   &   67,864 (3230)  &   50,556 (2406) \\
    $1$     &   42.02   &   68,703 (1635)  &   43,220 (1029) \\
    $2$     &   63.03   &   35,780 (568)  &   21,234 (337) \\
    $3$     &   84.04   &   25,533 (304)  &   14,284 (170) \\
    \midrule
    \multicolumn{4}{c}{Coreferences - split data} \\
    \midrule
    $d$    &   \#tokens    &   \multicolumn{2}{c}{Occurrences} \\
    &   &   All    &   Pronouns \\
    \midrule
    $0$     &   10.51     &   32,190 (3063)  &   24,328 (2315)   \\ 
    $1$     &   21.02   &   54,424 (2589)  &   37,966 (1806)   \\ 
    $2$     &   31.53   &   37,837 (1200)  &   23,732 (753)   \\ 
    $3$     &   42.04   &   22,529 (536)  &   14,035 (334) \\ 
    \midrule
    \multicolumn{4}{c}{Dependency trees} \\
    \midrule
    \multicolumn{3}{l}{\emph{Split} dependency} & Occurrences \\
    \midrule
    \multicolumn{2}{l}{subj or obj} &   &   41,065 \\
    \multicolumn{2}{l}{complement}  &   &   21,726 \\
    \multicolumn{2}{l}{modifier}    &   &   21,144 \\
    \multicolumn{2}{l}{any}         &   &   147,066  \\
    \bottomrule
    \end{tabular}
    \caption{Number of coreference antecedents at a given distance $d$ from the mention in the current sentence, for both original and split En$\rightarrow$Fr IWSLT17. In brackets, the same figure divided by the average number of tokens that the model has to attend to resolve the coreference (\#tokens). At the bottom, the number of sentences for which at least one syntactic dependency is split in two segments when using the split data. The percentage of examples that need context after splitting is 29.17\% if we consider pronominal coreferences only, 39.8\% if we consider all coreferences.}
    \label{tab:proofconcept}
\end{table}


\begin{table}[ht]
    \centering
    \small
    \begin{tabular}{lrrr}
    \toprule
    \multicolumn{4}{c}{Coreferences - original data} \\
    \midrule
    $d$    &   \#tokens    &   \multicolumn{2}{c}{Occurrences} \\
    &   &   All    &   Pronouns \\
    \midrule
    $0$     &   8.32   &   36,628 (4402)  &   27,179 (3267) \\
    $1$     &   16.64   &   60,204 (3618)  &   41,652 (2503) \\
    $2$     &   24.96   &   26,397 (1058)  &   16,142 (647) \\
    $3$     &   33.28   &   11,571 (348)  &   6,654 (200) \\
    \midrule
    \multicolumn{4}{c}{Coreferences - split data} \\
    \midrule
    $d$    &   \#tokens    &   \multicolumn{2}{c}{Occurrences} \\
    &   &   All    &   Pronouns \\
    \midrule
    $0$     &   4.16   &   13,322 (3202)  &   9,134 (2196)   \\
    $1$     &   8.32   &   46,227 (5556) &   34,104 (4099)   \\
    $2$     &   12.48   &   33,566 (2690)  &   22,676 (1817)   \\
    $3$     &   16.64   &   18,961 (1139)  &   12,248 (736)   \\
    \bottomrule
    \end{tabular}
    \caption{Same as in table~\ref{tab:proofconcept} for the Low Res En$\rightarrow$Ru.}
    \label{tab:proofofconcept_ru}
\end{table}

\begin{table*}[ht]
\centering
\small
\begin{tabular}{lcccccc}
\toprule
 & Total & $d=0$ & $d=1$ & $d=2$ & $d=3$ & $d>3$ \\
\midrule
\kz & 46.37 & \textbf{83.3} & 32.4 & 44.8 & 48.9 & 71.9 \\
\hdashline
\textit{K1} & 47.05 & 82.5 & 33.9 & 45.3 & 48.0 & 69.9 \\
\textit{K3} & 46.48 & 82.4 & 32.8 & 45.0 & 48.9 & 71.7 \\
\midrule
\textit{K1-\dr} & \textbf{60.21} & 81.1 & \textbf{56.5} & 44.9 & 48.7 & \textbf{73.3} \\
\textit{K3-\dr} & 56.22 & 81.7 & 46.8 & \textbf{55.2} & \textbf{56.2} & 72.4 \\
\midrule
Sample Size & 12000 & 2400 & 7075 & 1510 & 573 & 442 \\
Relative Size & 100.0\% & 20.0\% & 59.0\% & 12.6\% & 4.8\% & 3.7\% \\
\bottomrule
\end{tabular}
\caption{Accuracy(\%) of Low Res models on ContraPro En$\rightarrow$De by pronoun antecedent distance. The first column represents the weighted average, calculated on the basis of the sample size of each group.}
\label{tab:d}
\end{table*}


\begin{figure}
    \centering
    \includegraphics[scale=0.42]{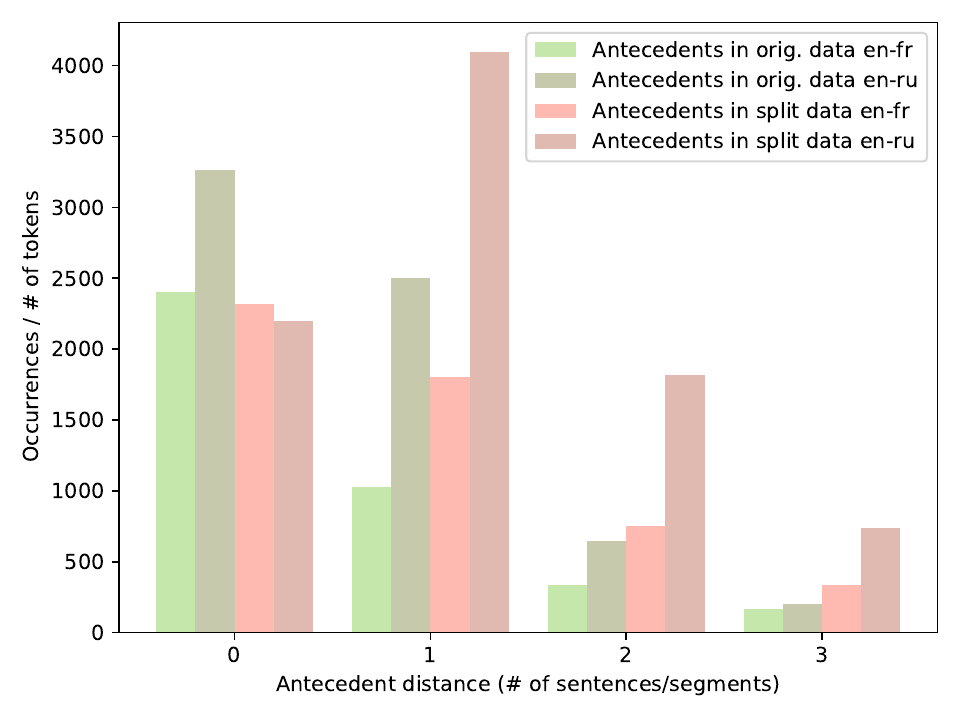}
    \caption{ En-Fr IWSLT vs Low Res En-Ru OpenSubtitles2018: comparison of the number of antecedents of anaphoric pronouns at a given distance in terms of sentences/segments, divided by the number of tokens that the model needs to attend for resolving the coreference. Since sentences are much shorter in En-Ru data (8.32 vs. 21.02 tokens on average), the density of discourse phenomena within the sentence is much higher.
    }
    \label{fig:normsplitcoref_comp}
\end{figure}


\section{Splitting methods}\label{app:methods}

We provide here some extra details on the splitting methods that have been proposed and tested. For full details, we refer to our implementation.

\textbf{Aligned-split.} As already mentioned, we use $\mathrm{wheresplit}=\mathrm{alignedsplit}(S^i, T^i, A^i)$ , which takes as input the word alignments $A^i$:
\begin{align*}
    A^i = \{(j,k)| S^i_j\text{ and }T^i_k\text{ are aligned}\},
\end{align*}

where $j=1,...,|S^i|$ and $k=1,...,|T^i|$ are the indices of the words belonging to $S^i$ and $T^i$, respectively. $\mathrm{alignedsplit}$ initially takes $m_S=\lfloor len(S^i)/2\rfloor$ and $m_T=\text{max}\{k\colon (j,k)\in A^i, j \le m_S\}$. Then, it checks whether this choice is not breaking apart two aligned words. Formally, it checks that:
\begin{gather}\label{eq:alignedsplit}
S^i_j\in S^{i,1} \wedge T^i_k \in T^{i,1} \text{ or } S^i_j\in S^{i,2} \wedge T^i_k \in T^{i,2}.
\end{gather}

If this condition is not encountered, it tries to split the sentence pairs in the neighbouring distance, where condition~(\ref{eq:alignedsplit}) is met. If the condition can not be met (e.g., because one of the two segments would be too short (<3 tokens)), $\mathrm{alignedsplit}$ falls back on $\mathrm{middlesplit}$.

\textbf{Synt-split.} In our implementation, the function $\mathrm{wheresplit}=\mathrm{syntsplit}(S^i, T^i, L^i)$ takes as input the coreference relation $L^i$ detected by CoreNLP 
on the source sentence $i$. If $L^i$ is not empty, it means that a relevant intra-sentential 
relation is present (in our experiments, we look at pronominal coreferences). In this case, the algorithm checks whether splitting in the middle ($m_S=\lfloor len(S^i)/2\rfloor$) allows to break $L^i$, i.e., to separate the two 
related tokens in different segments. If \textit{middle-split} 
does not achieve this goal, $m_S$ is set to the closest index from the middle that breaks the relation, except for the case in which breaking the 
relation would mean generating a too short segment (<3 tokens). In this case, the algorithm falls back to \textit{middle-split}.

\section{Impact of splitting}\label{app:impact}

In Table~\ref{tab:proofconcept}, we provide details on the syntactic features and the impact of splitting (with \textit{middle-split}) for En$\rightarrow$Fr IWSLT17, while Table~\ref{tab:proofofconcept_ru} shows the equivalent figures for the Low Resource subset of En$\rightarrow$Ru OpenSubtitles2018. A visual comparison of the two datasets is presented in Figure~\ref{fig:normsplitcoref_comp}. This complementary information confirms that the \textit{middle-split} method is an effective way to strengthen the contextual training signal and to facilitate its exploitation by context-aware NMT systems, in different text domains.

\begin{table*}[ht]
\centering
\small
\begin{tabular}{p{20mm}|p{19mm}p{20mm}|p{12mm}p{16mm}|p{12mm}p{16mm}}
\toprule
 & \multicolumn{2}{c}{\textbf{En$\rightarrow$Ru}} & \multicolumn{2}{c}{\textbf{En$\rightarrow$De}} & \multicolumn{2}{c}{\textbf{En$\rightarrow$Fr}} \\
 & \textbf{Low Res} & \textbf{Hig Res} & \textbf{Low Res} & \textbf{Hig Res} & \textbf{Low Res} & \textbf{Hig Res} \\
\midrule
Sentence-level training & OpenSubs2018 & OpenSubs2018 & WMT17 & WMT17 & WMT14 & WMT14 \\
\midrule
Context-aware training & 1/10th of \newline OpenSubs2018 & OpenSubs2018 & IWSLT17 & News-v12 \newline Europarl-v7 \newline IWSLT17 & IWSLT17 & News-v9 \newline Europarl-v7 \newline IWSLT17 \\
\midrule
Fine-tuning & - & - & - & IWSLT17 & - & IWSLT17 \\
\midrule
Test (BLEU) & OpenSubs2018 & OpenSubs2018 & IWSLT17 & IWSLT17 & IWSLT17 & IWSLT17 \\
\midrule
Contrastive test & EllipsisVP & EllipsisVP & ContraPro & ContraPro & ContraPro & ContraPro \\
\bottomrule
\end{tabular}
\caption{Summary of the datasets used at each stage of training and evaluation of the models.}
\label{tab:datarecap}
\end{table*}

\begin{table}[t!]
\centering
\small
\begin{tabular}{llccc}
\toprule
\textbf{Model} & \textbf{Setting} & \textbf{En$\rightarrow$Ru} & \textbf{En$\rightarrow$De} & \textbf{En$\rightarrow$Fr} \\
\midrule
\textit{K0} & - & 3.626 & 3.629 & 3.230 \\
\hdashline
\textit{K1} & Low Res & 3.599 & 3.617 & 3.216 \\
\textit{K3} & Low Res & 3.605 & 3.618 & 3.215 \\
\midrule
\textit{K1} & High Res & 3.596 & 3.617 & 3.210 \\
\textit{K3} & High Res & 3.597 & 3.617 & 3.211 \\
\midrule
\textit{K1-d\&r} & Low Res & 3.595 & 3.617 & 3.213 \\
\textit{K3-d\&r} & Low Res & 3.595 & 3.616 & 3.212 \\
\midrule
\textit{K1-d\&r} & High Res & 3.593 & 3.616 & 3.211 \\
\textit{K3-d\&r} & High Res & 3.592 & 3.615 & 3.211 \\
\bottomrule
\end{tabular}
\caption{Corresponding loss on development set for each reported test result with \textit{middle-split}.}
\label{tab:valid}
\end{table}

\begin{table}[t!]
\small
\centering
\begin{tabular}{cccc}
\toprule
$\mathbf{P_{len}}$ & \textbf{En$\rightarrow$Ru} & \textbf{En$\rightarrow$De} & \textbf{En$\rightarrow$Fr} \\
\midrule
0.6 & \textbf{31.76} & \textbf{32.80} & 44.47 \\
0.7 & 31.58 & 32.76 & 44.48 \\
0.8 & 31.47 & 32.72 & 44.50 \\
0.9 & 31.33 & 32.65 & 44.53 \\
1 & 31.23 & 32.64 & \textbf{44.59} \\
1.1 & 31.12 & 32.60 & \textbf{44.59} \\
1.2 & 31.06 & 32.57 & 44.58 \\
\bottomrule
\end{tabular}
\caption{Performance (BLEU) of \textit{K0} on the development set according to different values of length penalty.}
\label{tab:lenpen}
\end{table}

\begin{table*}[t!]
\small
\centering
\begin{tabular}{l|cc|cc}
\toprule
\textbf{} & \multicolumn{2}{c|}{\textbf{En$\rightarrow$De}} & \multicolumn{2}{c}{\textbf{En$\rightarrow$Fr}} \\
\textbf{Model} & \textbf{BLEU} & \textbf{ContraPro} & \textbf{BLEU} & \textbf{ContraPro} \\
\midrule
\textit{K0} & 32.97 (+0.00) & 46.37 (0.00) & 41.44 (-0.00) & 79.46 (0.00) \\
\textit{K1} & 33.06 (+0.06) & 46.7 (-0.35) & 41.75 (-0.12) & 79.05 (-0.19) \\
\textit{K3} & 32.73 (-0.13) & 46.21 (-0.27) & 41.47 (+0.15) & 79.24 (-1.29) \\
\midrule
\textit{K1-d\&r} & 33.1 (-0.34) & 47.6 (-12.61) & 41.64 (-0.14) & 78.94 (-5.12) \\
\textit{K3-d\&r} & 33.05 (-0.31) & 47.96 (-8.26) & 41.55 (-0.13) & 79.05 (-6.45) \\
\bottomrule
\end{tabular}
\caption{BLEU and accuracy results on ContraPro (and their changes) when the context provided to the model is inconsistent. All models are trained on the Low Resource setting.}
\label{tab:shuffles}
\end{table*}

\section{Experimental Setup}\label{app:setting}
\subsection{Data recap}\label{app:data}

We recap in Table~\ref{tab:datarecap} the datasets that we use at each stage of training and test. The sentence-level training concerns the baselines, whose parameters are also used to initialize the sentence-level encoder and decoder of the context-aware models ($\Theta_S$). Concerning En$\rightarrow$Ru, \citet{voita_when_2019} released two datasets extracted from OpenSubtitles2018: a document-level dataset of 1.5M sentences with context (document boundaries are available), and a sentence-level dataset of 6M sentences, which includes the sentences of the document-level dataset.

\subsection{Data preprocessing}\label{app:preprocessing}

The Opensubtitles2018 release by \citet{voita_when_2019} has been already pre-processed. Therefore, we only apply byte pair encoding \cite{sennrich_neural_2016} using 32k merge operations jointly for source and target languages.

The other datasets are tokenized with the Moses toolkit~\cite{koehn_moses_2007}, further cleaned by removing long sentences, and byte pair encoded using 32k merge operations jointly for source and target languages. While IWSLT provides document boundaries for TED subtitles, the WMT releases of New-Commentary and Europarl do not provide them. Therefore, a small fraction of sentences in the High Resource  setting will be paired with wrong context. However, we found the models to be robust against occasional random context (see also \citet{voita_context-aware_2018} and \citet{muller_large-scale_2018}). In order to make the models correctly learn how to translate headlines (the first line in a document), we need to have headlines in the training set. As such, we set artificial document boundaries in News-Commentary and Europarl, following the average document length of TED talks.

\subsection{Training and evaluation}\label{app:training}
All models are implemented in \textit{fairseq}~\cite{ott_fairseq_2019}. After having pre-trained the baseline on 4 Tesla V100 for 200k steps, we train all models on a single Quadro RTX 6000, with a fixed batch size of approximately 16k tokens,\footnote{The optimizer update is delayed to simulate  16k tokens.} as it has been shown that Transformers need a large batch size for achieving the best performance \citep{popel_training_2018}. We stop training after 5 consecutive non-improving validation steps (in terms of loss on dev).  Corresponding validation performance for each reported test result with \textit{middle-split} are reported in Table~\ref{tab:valid}. We train models with the optimizer configuration and learning  rate (LR)  schedule  described in \citet{vaswani_attention_2017}. The maximum LR is 0.0007 for baselines on En$\rightarrow$Ru/De, 0.001 for models on En$\rightarrow$De/Fr low resource settings, and 0.0005 for all the others. In the En$\rightarrow$De/Fr High Resource setting, contextual-parameters are finetuned on IWSLT17 with an initial LR of 0.0002 that shrinks by a factor of 0.99 at every epoch. We use label smoothing with an epsilon value of 0.1~\citep{Pereyra_regularizing_2017} for all settings. Since the sentence-level parameters are pre-trained on a large amount of parallel data (WMT), the models are pretty robust to generalization, and dropout can be set to 0.1, which gave the best results for the non-contextual baseline \kz. At inference time, we use beam search with a beam of 4 for all models. We adopt a length penalty ($P_{len}$) of 0.6 for all models ($P_{len}<1$ favors shorter sentences), with the exception of En$\rightarrow$Fr models, to which we assign $P_{len}=1$. The LR for training was searched in $\{0.001,0.0007,0.0005,0.0002\}$). The LR achieving the best loss on the validation set after convergence was selected. $P_{len}$ was searched in $\{0.6,0.7,0.8,0.9,1.0,1.1,1.2\}$ for \textit{K0} only (see Table~\ref{tab:lenpen}). The length penalty resulting in the best BLEU score on the validation set was then used for all models within the same language pair. The other hyperparameters were set according to the relevant literature~\citep{vaswani_attention_2017, popel_training_2018, voita_when_2019, lopes_document-level_2020}.

\section{Results Analysis}\label{app:bydistance}

\subsection{Accuracy by antecedent distance}\label{app:d}

Here, we want to investigate more in details the performance of the proposed approach on the accuracy of ambiguous pronoun translation. We report in Table~\ref{tab:d} the accuracy on En$\rightarrow$De ContraPro, detailed by varying antecedent distance. We notice that all the improvements achieved by \textit{-d\&r} models are related to those pronouns whose antecedent is in the context ($d\geq1$), which is in line with the expectations of context-aware models exploiting context for disambiguation. \textit{K1-d\&r} is very strong in translating pronouns with antecedent distance $d=1$, surpassing \kz{} and \textit{K1} baselines by 22+ points of accuracy. Similarly, \textit{K3-d\&r} surpasses baselines by a large margin on $0\leq d\leq 3$, beating all the other models on $d=2,3$, as expected. We notice however that \textit{K3-d\&r} lacks behind \textit{K1-d\&r} on $d=1$. On one side, this could be explained by the fact that \textit{K1-d\&r} is more specialized at modeling a single past sentence. On the other side, we also notice that the hierarchical context-encoding architecture by~\citet{miculicich_document-level_2018}, at the core of \textit{K3}, is not aware of the distance of the context sentences that are encoded. Hence, we believe that \textit{K3-d\&r} might perform worse on $d=1$ than \textit{K1-d\&r} because it gives the same importance to further away context ($d=2,3$). Since pronouns with antecedent distance $d=1$ are the most frequent in the test set, \textit{K1-d\&r} has the highest average result (reported in ``Total"). It has to be noticed also that {\kt} is more affected by the challenge of sparsity than \ko, since it has to spot relevant context among 3x more tokens. This might be the reason why {\kt} starts beating {\ko} only when the training setting is the most favorable to context-aware learning: with \dr pre-training plus high resources.


\subsection{Ablation: shuffling context}

We want to verify that the proposed approach improves learning by making the context-aware model to rely on its modeling of the context. Table~\ref{tab:shuffles} shows the performance of models trained on Low Res, when the evaluation is undertaken by randomly shuffling the context of every sentence with other sentences from the same dataset  (c.f. \citet{scherrer_analysing_2019}). In brackets, the delta w.r.t. the results with consistent context presented in the  main table of the paper. A random context is inconsistent with the current sentence in many cases, and thus misleading for a context-aware system. Indeed,  -\dr models display a significant drop in accuracy when they are evaluated with inconsistent context, which confirms that they rely on context information to achieve the improvement in pronoun translations. Nonetheless, the same models prove to be robust against being shown a random context as they obtain a similar performance to \kz. In other words, the splitting method does not produce models that are over reliant on context. This robustness is confirmed by BLEU: the average translation quality is very slightly affected by the shuffling. The changes are so small that are probably negligible. This results also show once again that BLEU is ill-equipped to measuring improvements in document-level translation.


\end{document}